\documentclass[10pt,twocolumn,letterpaper]{article}

\usepackage[final]{cvpr}      

\newcommand{\name}{\textbf{TraceR1}}

\usepackage{xcolor}
\definecolor{cvprblue}{rgb}{0.21,0.49,0.74}
\definecolor{cvprlightgray}{rgb}{0.45,0.45,0.45}
\definecolor{cvprcyan}{rgb}{0.15,0.65,0.80}
\definecolor{cvprpurple}{rgb}{0.54, 0.35, 0.95}
\definecolor{cvprgreen}{rgb}{0.23, 0.70, 0.35}
\definecolor{lightblue}{RGB}{255,240,245}
\definecolor{lightgreen}{RGB}{255,255,240}
\usepackage{fontawesome5}
\usepackage{tcolorbox}
\usepackage{enumitem}
\usepackage{booktabs}      
\usepackage{multirow}       
\usepackage{graphicx}       
\usepackage{array}       
\usepackage{subcaption}
\usepackage{rotating} 
\usepackage{wrapfig}   
\usepackage{graphicx}
\usepackage{listings}
\usepackage{bbm}

\lstdefinestyle{LLMQuery}{
    basicstyle=\ttfamily\small,
    backgroundcolor=\color{gray!5},
    frame=single,
    breaklines=true,
    columns=fullflexible,
    keywordstyle=\color{blue}\bfseries,
    commentstyle=\color{gray}\itshape,
    showstringspaces=false,
}
\usepackage{colortbl}
\tcbuselibrary{skins}
\newtcolorbox{promptbox}[1]{%
  colback=blue!5,
  colframe=blue!30,
  fonttitle=\bfseries,
  title={#1},
  boxrule=0.5pt,
  sharp corners
}
\usepackage{listings}
\lstdefinelanguage{json}{
  basicstyle=\normalfont\ttfamily,
  numbers=left,
  numberstyle=\scriptsize,
  stepnumber=1,
  numbersep=8pt,
  showstringspaces=false,
  breaklines=true,
  frame=lines,
  backgroundcolor=\color{white},
  literate={"}{{\texttt{"}}}1 {,}{{\texttt{,}}}1 {:}{{\texttt{:}}}1
           {0}{{\texttt{0}}}1 {1}{{\texttt{1}}}1 {2}{{\texttt{2}}}1 {3}{{\texttt{3}}}1
           {4}{{\texttt{4}}}1 {5}{{\texttt{5}}}1 {6}{{\texttt{6}}}1 {7}{{\texttt{7}}}1
           {8}{{\texttt{8}}}1 {9}{{\texttt{9}}}1,
}
\usepackage{adjustbox}
\usepackage[pagebackref,breaklinks,colorlinks,allcolors=cvprblue]{hyperref}

\def\paperID{****} 
\def\confName{CVPR}
\def\confYear{2026}

\title{Anticipatory Planning for Multimodal AI Agents}

\author{Yongyuan Liang\textsuperscript{1} \ \ \ Shijie Zhou\textsuperscript{4} \ \ \ Yu Gu\textsuperscript{2} \ \ \ Hao Tan\textsuperscript{3} \ \ \ Gang Wu\textsuperscript{3} \\
Franck Dernoncourt\textsuperscript{3} \ \ \ Jihyung Kil\textsuperscript{3} \ \ \ Ryan A. Rossi\textsuperscript{3} \ \ \ Ruiyi Zhang\textsuperscript{3} \\
$^{1}$\small{University of Maryland, College Park,} $^{2}$\small{The Ohio State University,} \\
$^{3}$\small{Adobe Research,}
$^{4}$\small{State University of New York at Buffalo}
\\
}

\begin{document}
\maketitle
\begin{abstract}
Recent advances in multimodal agents have improved computer-use interaction and tool-usage, yet most existing systems remain reactive, optimizing actions in isolation without reasoning about future states or long-term goals. This limits planning coherence and prevents agents from reliably solving high-level, multi-step tasks.
We introduce \name, a two-stage reinforcement learning framework that explicitly trains anticipatory reasoning by forecasting short-horizon trajectories before execution. The first stage performs trajectory-level reinforcement learning with rewards that enforce global consistency across predicted action sequences. The second stage applies grounded reinforcement fine-tuning, using execution feedback from frozen tool agents to refine step-level accuracy and executability.
\name{} is evaluated across seven benchmarks, covering online computer-use, offline computer-use benchmarks, and multimodal tool-use reasoning tasks, where it achieves substantial improvements in planning stability, execution robustness, and generalization over reactive and single-stage baselines. These results show that anticipatory trajectory reasoning is a key principle for building multimodal agents that can reason, plan, and act effectively in complex real-world environments.
\end{abstract}    
\section{Introduction}
\label{sec:intro}
\begin{figure*}[t]
    \centering
    \includegraphics[width=0.99\linewidth]{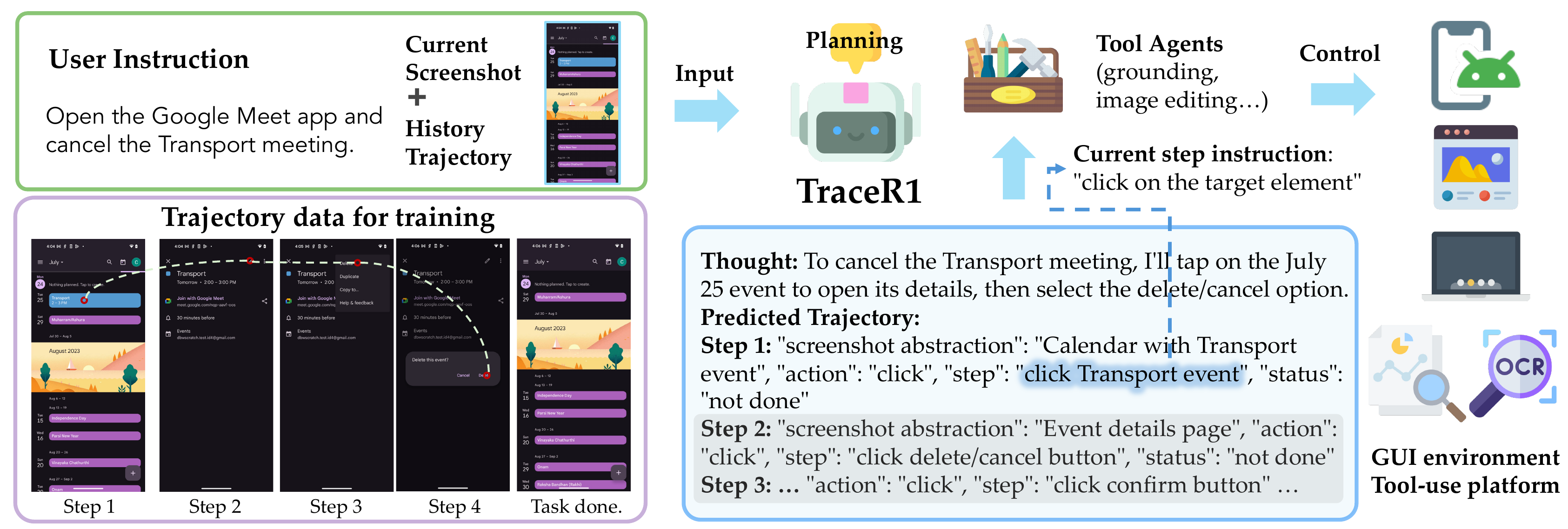}
    \vspace{-1mm}
    \caption{\textbf{\name{} overview:} anticipatory planning and grounded execution. 
    The planning model takes a user instruction together with the current screenshot and interaction history, predicts a trajectory of future actions, \emph{where only \textcolor{cvprblue}{the first predicted step} is executed by the tool agent while \textcolor{cvprlightgray}{later steps are unexecuted lookahead predictions}}, and generates step-level instructions for tool agents to complete tasks across diverse GUI environments and tool-use platforms.}
    \label{fig:overview}
\end{figure*}

Building intelligent agents that can plan and act over long horizons has long been a central goal in the era of large language and multimodal agents~\citep{claude4,openai2025o3,Agent-S2,yang2025qwen3}.
Recent advances in multimodal autonomous agents have shown impressive capabilities in GUI interaction~\citep{nguyen2025gui}, embodied control~\citep{yang2025magma,zheng2024tracevla}, and tool-use reasoning~\citep{su2025thinking}.
However, despite their strong reasoning priors, most existing multimodal agents remain fundamentally \emph{reactive}: they decide the next action based only on the current observation, focusing on immediate perception without anticipating the long-term consequences of their decisions. Without anticipatory reasoning, agents tend to fail in multi-step environments where actions have delayed and compounding effects, causing them to gradually diverge from the intended task.

To develop multimodal agentic models capable of looking ahead, two major directions have been explored.
Model-free reinforcement learning (RL)~\citep{luo2025gui, liu2025infigui, tang2025magicgui, wang2025ui} trains agents through step-level action correctness and designed rewards for subgoals or sparse final outcomes.
Model-based planning~\citep{gu2024your, luo2025vimo, chen2025planning, gao2025uishift}, in turn, equips agents with a world model that simulates future action sequences and evolving environment states, enabling them to reason about possible outcomes before acting.
Yet both approaches face fundamental obstacles: constructing world models over visually rich and interactive environments is notoriously difficult, and defining reasoning-oriented rewards that generalize across diverse and open-ended tasks remains an open challenge.
This raises the question: \textit{how can we efficiently train multimodal agents to develop adaptive anticipatory reasoning for complex, long-horizon tasks?}

We address this challenge by introducing \name, a two-stage RL framework designed to combine long-horizon trajectory reasoning with grounded execution refinement.
In the first stage, \textit{anticipatory trajectory optimization}, the model performs
trajectory-level RL on large-scale agent trajectories. The rewards evaluate global
consistency between the predicted and reference action sequences, encouraging coherent
planning and anticipatory reasoning over multiple future steps. In the second stage,
\textit{grounded reinforcement fine-tuning}, the model is refined using step-level
executable feedback from tool agents. Grounded rewards, such as coordinate
accuracy and answer correctness, improve precision and ensure that each predicted step
remains feasible within the environment.
This two-stage structure resembles how humans plan: \textit{anticipating several steps ahead and then refining the immediate action based on feedback.}

By explicitly modeling future dependencies while grounding each action in executable feedback, \name{} provides a general training recipe for GUI environments, tool-use systems, and multimodal reasoning tasks.
Empirically, it achieves substantial improvements in both planning stability and execution robustness, attaining planning capability comparable to proprietary systems, significantly outperforming open-source baselines on long-horizon GUI benchmarks such as OSWorld-Verified~\citep{osworld_verified} and AndroidWorld~\citep{li2024effects}, and demonstrating strong reasoning and execution reliability on general tool-use benchmarks including GAIA~\citep{mialon2023gaia} and GTA~\citep{wang2024gta}.
These results highlight anticipatory trajectory reasoning as a key step toward building planning agents that can reason and plan with foresight while advancing long-horizon goals in complex, real-world environments.

In summary, the main contributions of this work are:

\begin{itemize}
    \item We introduce \name, a unified framework for anticipatory planning that forecasts trajectories of future actions and step-level instructions, enabling long-horizon reasoning and foresight beyond reactive decision making.

    \item We develop a two-stage reinforcement learning paradigm that first performs trajectory-level optimization to learn globally coherent plans and then applies grounded reinforcement fine-tuning with executable feedback, bridging high-level reasoning and low-level precision across GUI and tool-use environments.
    
    \item We conduct comprehensive evaluations across $7$ GUI and multimodal tool-use reasoning benchmarks, demonstrating substantial improvements in planning stability, execution robustness, and generalization, achieving performance comparable to proprietary systems and surpassing open-source baselines.

\end{itemize}
\begin{figure*}[t]
    \centering
    \includegraphics[width=0.99\linewidth]{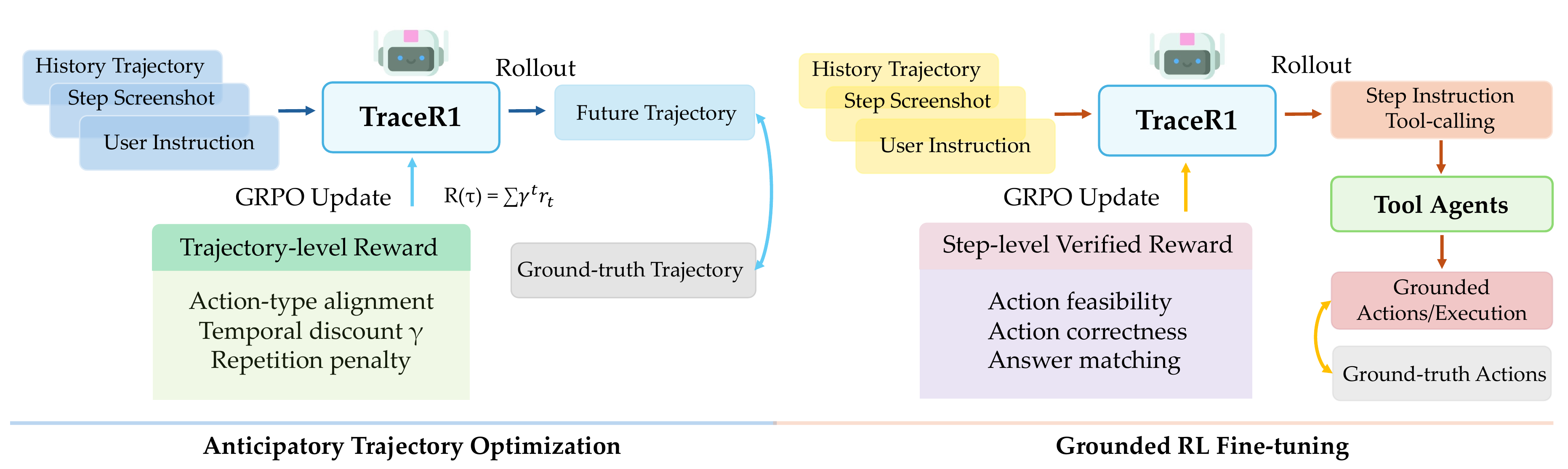}
    \vspace{-1mm}
    \caption{\textbf{Two-stage training framework of \name.} 
    Stage~$1$ performs anticipatory trajectory optimization using trajectory-level
    alignment rewards, while Stage~$2$ applies grounded RL fine-tuning with
    step-level rewards derived from tool-agent execution feedback.}
    \label{fig:training}
\end{figure*}

\section{Related Work}
\label{sec:related}

\subsection{Planning-Oriented GUI Agents}
\noindent
\textbf{Agent frameworks.}
Recent GUI agent frameworks increasingly emphasize structured planning pipelines that combine reasoning modules with grounding and execution components.
Agent systems such as Aria-UI~\citep{yang2025aria}, UGround~\citep{gou2024navigating}, SeeClick~\citep{cheng2024seeclick}, Jedi~\citep{xie2025scaling}, Agent S/S2~\citep{agashe2024agent,Agent-S2}, and GTA1~\citep{yang2025gta1} all follow this paradigm, typically employing powerful API-based proprietary models such as o3~\citep{openai2025o3} or Claude 4~\citep{claude4} as planners to generate high-level action proposals, while domain-specific modules handle grounding and execution on GUI interfaces.
These frameworks have demonstrated impressive multi-step reasoning and cross-platform control, yet their progress largely depends on the underlying proprietary planners rather than improving the agent’s intrinsic planning capability.
They emphasize precise action execution based on instructions over trajectory-level planning, whereas our work directly trains large multimodal models to acquire anticipatory planning through RL.

\noindent
\textbf{Generalist agents.}
A parallel line of work builds generalist agents on top of large vision–language models~\citep{openai2025o3, claude4, yang2025qwen3}, extending them to computer use, GUI control, and a broad range of agentic tasks.
Research efforts such as UI-TARS~\citep{qin2025ui,wang2025ui}, Magma~\citep{yang2025magma}, and OpenCUA~\citep{wang2025opencua} develop unified pipelines for interactive control and reasoning across diverse GUI environments, while models including SeeAct~\citep{zheng2024gpt}, CogAgent~\citep{hong2024cogagent}, and OS-ATLAS~\citep{wu2024atlas} emphasize perception–reasoning integration for interface understanding and task decomposition.
Recent R1-style approaches further incorporate reinforcement signals to enhance agent reasoning in GUI settings~\citep{luo2025gui,liu2025infigui,lu2025ui}.
Unlike these methods, which still rely on grounding supervision and emphasize precise action execution during training, our approach focuses purely on planning and introduces a more general training framework that strengthens a multimodal agent’s ability to plan, comprehend, and anticipate future states.

\subsection{Tool-Usage Multimodal Agents}
The ability to use external tools is a defining aspect of intelligent multimodal agents, allowing them to perform complex, visually grounded tasks beyond direct perception and reasoning.
One line of research enhances this capability through large-scale multimodal instruction tuning, where models learn tool selection and composition from synthetic or curated trajectories~\citep{yang2023gpt4tools,liu2024llava,wang2025mllm}.
Another line builds end-to-end architectures that couple vision–language models with real executable tools or interactive environments, enabling stepwise control and adaptive reasoning~\citep{gao2024multi,wang2024jarvis,zhang2025appagent,zhang2024you,chen2023llava}.
These methods substantially improve tool invocation and multimodal integration but primarily emphasize execution reliability or reactive coordination.
In contrast, our approach focuses on strengthening the agent’s planning capability by training models to anticipate and organize future tool-use behaviors, using grounded feedback solely for execution validation rather than as the primary learning signal, thereby enabling more effective and deliberate tool-use reasoning.
\section{Methodology}
\label{sec:method}
\name{} is trained with a two-stage RL framework designed to
enable anticipatory multimodal planning. In this section, we introduce the agent formulation, followed by the two training stages.

\noindent
\textbf{Problem Formulation.}
At step $t$, the agent receives the current observation $s_t$ and predicts an
action $a_t$ and step instruction $g_t$. It also conditions on a compact
interaction history $\tau_{1:t-1} = \{(\phi(s_i), a_i)\}_{i=\max(1,\, t-K)}^{t-1}$,
where $\phi(s_i)$ is an abstracted summary of the past observation rather than a
raw screenshot. The predicted action is executed by a tool agent, and
the resulting observation becomes the next state. This $K$-step truncated history
provides lightweight temporal context while avoiding redundancy.

To train such an agent, we adopt the two-stage reinforcement learning framework shown in
Figure~\ref{fig:training}, which integrates long-horizon trajectory alignment with
grounded execution refinement.
Stage~$1$ performs \emph{anticipatory trajectory optimization}, aligning predicted and reference trajectories via trajectory-level rewards that encourage globally consistent plans.
Stage~$2$ performs \emph{grounded reinforcement fine-tuning}, incorporating feedback from tool agents to refine step-level accuracy and execution feasibility.

\subsection{Anticipatory Trajectory Optimization}
Supervised fine-tuning (SFT) on next-step predictions enables an agent to imitate local behaviors but struggles to capture long-term dependencies. Even when trained on full trajectories, SFT optimizes token- or step-level likelihoods under teacher forcing, neglecting global consistency and failing to penalize redundant or unstable rollouts. 

To address these limitations, \name{} performs trajectory-level RL that aligns
predicted and reference trajectories within a bounded horizon, encouraging the
agent to reason several steps ahead before acting. Each training sample contains a
user instruction $u$, the current observation $s_t$, and a reference trajectory
$\tau^{*} = \{(a_1^{*}, g_1^{*}), \ldots, (a_T^{*}, g_T^{*})\}$, where $a_t^{*}$ and
$g_t^{*}$ denotes the ground-truth action type and step instruction. Conditioned on
$(u, s_t, \tau_{1:t-1})$, the model predicts a future trajectory
$\hat{\tau}_{t:T} = \{(\hat{a}_t, \hat{g}_t), \ldots, (\hat{a}_T, \hat{g}_T)\}$,
which is optimized via trajectory-level alignment rewards.

Training aligns the predicted and reference trajectories through a discounted trajectory-level reward:
\begin{equation}
R(\hat{\tau}, \tau^{*}) = \sum_{t=1}^{T} \gamma^{\,t-1} r_t ,
\end{equation}
where $\gamma \in (0,1)$ is the temporal discount factor and $r_t$ is the per-step alignment reward:
\begin{equation}
r_t =
\lambda_{\text{align}} \, \text{sim}(\hat{a}_t , a_t^{*})
\;-\;
\lambda_{\text{rep}} \, \text{rep}(\hat{a}_{1:t}),
\end{equation}
where $\text{sim}(\cdot,\cdot)$ measures the alignment between the predicted
operation $\hat{a}_t$ and the reference $a_t^{*}$ (GUI action type or tool call),
and $\text{rep}(\hat{a}_{1:t})$ penalizes repeated or cyclic actions within the
trajectory prefix. $\lambda_{\text{align}}$ and $\lambda_{\text{rep}}$ control the strengths of
action alignment and loop-prevention.

The policy $\pi_{\theta}$ is optimized using the group-relative policy optimization (GRPO) objective~\cite{guo2025deepseek}:
\begin{equation}
\begin{aligned}
\nabla_{\theta} J(\theta)
&= \mathbb{E}_{\hat{\tau}}\!\Big[
\hat{A}(\hat{\tau}, \tau^{*}) \nabla_{\theta} \log \pi_{\theta}(\hat{\tau}|u, s_t, \tau_{1:t-1}) 
\Big],
\end{aligned}
\end{equation}
where $\hat{A}(\hat{\tau}, \tau^{*})$ is the normalized group-relative advantage computed from $R(\hat{\tau}, \tau^{*})$.
Through this stage, the model learns to anticipate long-term effects before execution, improving the global coherence of multi-step plans.
\subsection{Grounded Reinforcement Fine-tuning}

While trajectory-level optimization promotes consistency across steps, accurate control still depends on grounding—ensuring that each predicted action leads to correct and feasible execution within the environment or tool interface.
Given $(u, s_t, \tau_{1:t-1})$, the model outputs $(\hat{a}_t, \hat{g}_t)$, which are executed by a frozen tool agent (e.g., GUI executor or callable tool modules).
The resulting outputs are compared with ground-truth responses to compute a step-level grounded reward $r_t^{\text{G}}$:
\begin{equation}
r_t^{\text{G}} =
\begin{cases}
\mathbbm{1}[\text{coord match}], & \text{for grounding steps},\\[4pt]
\mathbbm{1}[\text{answer match}], & \text{for tool-calling steps}.
\end{cases}
\end{equation} 
Here, $\mathbbm{1}[\text{grounding step}]$ and $\mathbbm{1}[\text{tool-calling step}]$ select the appropriate reward type for different tasks. This formulation applies coordinate
matching for GUI grounding steps and answer matching for tool-calling steps.

Grounded fine-tuning follows the same GRPO update rule as Stage~$1$, replacing the trajectory-level reward with the grounded step reward:
\begin{equation}
\begin{aligned}
\nabla_{\theta} J_{\text{G}}(\theta)
&= \mathbb{E}\!\Big[
\hat{A}(r_t^{G}) \nabla_{\theta} \log \pi_{\theta}(\hat{a}_t, \hat{g}_t \mid u, s_t, \tau_{1:t-1}) 
\Big].
\end{aligned}
\end{equation}
This stage refines execution precision and robustness while preserving the anticipatory structure learned during trajectory alignment.

\paragraph{Training Pipeline.}
In practice, both stages are trained with large-scale multimodal agent trajectory datasets, where each step, along with its subsequent action sequence, forms a training instance.
Stage~$1$ uses the full reference trajectories: for each step, the model predicts a short-horizon rollout, and the trajectory-level reward measures how well the entire predicted future sequence matches the ground-truth continuation, without executing any action.
Stage~$2$ uses the same per-step multi-step prediction setup, but only the first predicted action is executed by a frozen tool agent. The tool’s output (e.g., click coordinates or textual response) is compared with the corresponding ground-truth action or answer to compute a grounded reward.
This offline-grounded setup enables the model to learn anticipatory planning while using offline trajectories as the source of both trajectory-level and execution-level supervision.

\subsection{Inference with Anticipatory Planning}
\label{subsec:inference}
At inference time, \name{} operates in a plan–act loop.
Given the current observation, it predicts a multi-step future trajectory $\hat{\tau}_{t:T}$, executes only the first action via the tool agent, receives the updated environment feedback, and re-plans for the next step.
This iterative foresight mechanism allows the model to anticipate long-term outcomes while maintaining execution stability across diverse tool-use scenarios.
\begin{table*}[t]
\centering
\small
\setlength{\tabcolsep}{8pt}
\begin{tabular}{lcccc}
\toprule
\textbf{Agent Model} & \textbf{Params} & \textbf{AndroidWorld} & \textbf{OSWorld-Verified} \\
\midrule
\rowcolor{cvprcyan!10}
\multicolumn{4}{l}{\textit{Proprietary Models}} \\
o3~\citep{openai2025o3} & - & - & 23.0 \\
OpenAI CUA-o3~\citep{openai2025o3} & - & 52.5 & 38.1 \\
Seed1.5-VL~\citep{guo2025seed1} & - & 62.1 & 36.7 \\
Claude 4 Sonnet~\citep{claude4} & - & - & 41.4 \\
Claude 4.5 Sonnet~\citep{claude4} & - & - & 62.9 \\
UI-TARS-1.5~\citep{qin2025ui} & - & 64.2 & 41.8 \\
UI-TARS-2~\citep{wang2025ui} & - & 73.3 & 53.1 \\
\midrule
\rowcolor{cvprcyan!10}
\multicolumn{4}{l}{\textit{Agent System (w/ Proprietary Models)}} \\
Jedi-7B w/ o3~\citep{xie2025scaling} & 7B w/ - & - & 50.2 \\
Agent S2 w/ GPT-5~\citep{Agent-S2} & 7B w/ - & - & 48.8 \\
Agent S2 w/ Claude 3.7 Sonnet~\citep{Agent-S2} & 7B w/ - & 54.3 & - \\
Agent S2.5 w/ o3~\citep{Agent-S2} & 7B w/ - & - & 56.0 \\
GTA1-7B w/ o3~\citep{yang2025gta1} & 7B w/- & - & 53.1 \\
GTA1-32B w/ o3~\citep{yang2025gta1} & 32B w/- & - & 55.4 \\
UI-TARS-1.5-7B w/ GPT-4.1~\citep{qin2025ui} & 7B w/ - & - & 31.6 \\
Qwen3-VL-32B-Thinking w/ GPT-4.1~\citep{yang2025qwen3} & 32B w/ - & - & 43.2 \\
\midrule
\rowcolor{cvprcyan!10}
\multicolumn{4}{l}{\textit{Open-Source Models}} \\
Qwen2.5-vl-72b~\citep{bai2025qwen2} & 72B & 35.0 & 5.0 \\
UI-TARS-1.5-7B~\citep{qin2025ui} & 7B & - & 27.4 \\
UI-TARS-72B-DPO~\citep{qin2025ui} & 72B & - & 27.1 \\
OpenCUA-7B~\citep{wang2025opencua} & 7B & - & 26.6 \\
OpenCUA-32B~\citep{wang2025opencua} & 32B & - & 34.8 \\
Qwen3-VL-8B-Thinking & 8B~\citep{yang2025qwen3} & 50.0 & 33.9 \\
Qwen3-VL-32B-Thinking~\citep{yang2025qwen3} & 32B & 61.4 & 35.6 \\
Qwen3-VL-235B-A22B-Thinking~\citep{yang2025qwen3} & 236B & - & 38.1 \\
\midrule
\rowcolor{cvprcyan!10}
\multicolumn{4}{l}{\textit{\name}} \\
UI-TARS-1.5-7B w/ \textbf{Ours} & 7B w/ 8B & - & 30.9\\
Qwen3-VL-32B-Thinking w/ \textbf{Ours} & 32B w/ 8B & \textbf{64.8} & \textbf{41.2} \\
\bottomrule
\end{tabular}
\vspace{-1mm}
\caption{\textbf{Success rate (\%) on AndroidWorld and OSWorld-Verified.} 
OSWorld-Verified is evaluated under a 100-step maximum setting. }
\label{tab:gui-online}
\end{table*}

\begin{promptbox}{Preliminary Finding}
\itshape
  \textbf{\name{} takes one step while seeing several ahead.}
   Anticipatory reasoning allows the planning agent to account for long-term dependencies and downstream consequences,
  leading to more globally consistent and accurate planning decisions.
\end{promptbox}
\section{Experiment}
\label{sec:exp}
To comprehensively evaluate \name, we focus on GUI agent benchmarks that assess agents’ planning and interaction abilities across multiple platforms, and on tool-use benchmarks that examine general multimodal reasoning and problem-solving capability.

\subsection{Setup}

\noindent
\textbf{Implementation details.} 
Our model is initialized from Qwen3-VL-8B-Thinking~\citep{yang2025qwen3} and trained using the EasyR1 framework~\citep{zheng2025easyr1}. 
The training covers both GUI and multimodal tool-use datasets.

For GUI tasks, Stage~$1$ pretraining uses trajectory datasets from AgentNet~\citep{wang2025opencua}, 
AndroidControl~\citep{li2024effects}, 
GUI-Odyssey~\citep{lu2025guiodyssey}, 
Multimodal-Mind2Web~\citep{deng2023mind2web}, 
and AgentTrek~\citep{xu2024agenttrek},
adopting the structured action space defined in~\citep{qin2025ui} for unified cross-platform control.
Stage~$2$ performs grounded RFT using datasets from different GUI platforms with corresponding tool agents, including 
UI-TARS-7B~\citep{qin2025ui}, 
UI-TARS-1.5-7B~\citep{qin2025ui}, 
and Qwen3-VL-32B-Thinking~\citep{yang2025qwen3}.

For multimodal tool-use, Stage~$1$ leverages the tool-use trajectory dataset from~\citep{gao2024multi} following their standardized toolbox interface.
Stage~$2$ grounded RFT is then conducted with real-executable tools provided by the T3-Agent toolbox~\citep{gao2024multi}. Refer to Supplementary Material for more details.

\noindent
\noindent
\textbf{Benchmarks.} 
We evaluate \name{} across $7$ benchmarks that collectively measure GUI task execution and multimodal tool-usage reasoning.

The GUI benchmarks include both online agent capability evaluation, featuring dynamic and interactive environments simulating real-world scenarios, and offline evaluation, which measures agent performance in static, pre-defined settings. 
The online benchmarks comprise OSWorld-Verified~\citep{xie2024osworld}, which examines long-horizon desktop operations, and AndroidWorld~\citep{rawles2024androidworld}, which tests mobile task completion on a live Android emulator with 116 tasks across 20 applications; both use task success rate as the evaluation metric. 
The offline benchmarks consist of AndroidControl-High~\citep{li2024effects}, GUI-Odyssey~\citep{lu2025guiodyssey}, and Multimodal-Mind2Web~\citep{deng2023mind2web}, all evaluated by step success rate. 
AndroidControl-High targets high-level mobile execution, GUI-Odyssey focuses on cross-application navigation with 203 tasks spanning six apps, and Multimodal-Mind2Web extends Mind2Web to test generalization across cross-task, cross-website, and cross-domain settings. 

The tool-use and reasoning benchmarks include GTA~\citep{wang2024gta} and GAIA~\citep{mialon2023gaia}. 
GTA contains 229 tasks with 252 images requiring two to eight reasoning steps, evaluating perception, operation, logic, and creativity on visual data, while GAIA consists of 446 tasks involving 109 files (PPTX, PDF, XLSX, etc.) grouped into three difficulty levels, assessing document understanding, web reasoning, and answer summarization.
\begin{table*}[t]
\centering
\small
\setlength{\tabcolsep}{6pt}
\resizebox{\textwidth}{!}{
\begin{tabular}{lcccc}
\toprule
\textbf{Agent Model} & \textbf{Params} & \textbf{AndroidControl-High} & \textbf{GUI Odyssey} & \textbf{Multimodal-Mind2Web} \\
\midrule

\rowcolor{cvprpurple!10}
\multicolumn{5}{l}{\textit{Proprietary Models}} \\
GPT-4o~\citep{openai2025o3} & - &  21.2 &  5.4 & 4.3\\
Claude-computer-use~\citep{claude4} & - & 12.5 & 3.1 & 52.5 \\
\midrule

\rowcolor{cvprpurple!10}
\multicolumn{5}{l}{\textit{Agent System (w/ Proprietary Models)}} \\
OmniParser-v2.0 w/ GPT-4o~\citep{lu2024omniparser}  & - &  58.8 &  62.0 & 41.3\\
UI-TARS-7B w/ GPT-4.1~\citep{qin2025ui} & 7B w/ - &  74.8 &  89.1 & 66.0 \\
\midrule

\rowcolor{cvprpurple!10}
\multicolumn{5}{l}{\textit{Open-Source Models}} \\
OS-Atlas-4B~\citep{wu2024atlas} & 4B & 22.7 & 56.4 & -\\
OS-Atlas-7B~\citep{wu2024atlas} & 7B & 29.8 &   62.0 & -\\
QwenVL2.5-3B~\citep{bai2025qwen2} & 3B & 38.9 & 50.9 & -\\
QwenVL2.5-7B~\citep{bai2025qwen2} & 7B & 47.1 & 54.5 & -\\
GUI-R1-3B~\citep{luo2025gui} & 3B & 46.5 & 41.3 & -\\
GUI-R1-7B~\citep{luo2025gui} & 7B & 51.7 & 38.8 & -\\
InfiGUI-R1-3B~\citep{liu2025infigui} & 3B & 71.1 & -  & - \\
UI-TARS-2B~\citep{qin2025ui}  & 7B & 68.9 & 83.4 & 53.1\\
UI-TARS-7B~\citep{qin2025ui}  & 7B & 72.5 & 87.0 & 63.1\\
UI-TARS-32B~\citep{qin2025ui}  & 7B & 74.7 & 88.6 & 64.7\\
\midrule

\rowcolor{cvprpurple!10}
\multicolumn{5}{l}{\textit{\textbf{TraceR1}}} \\
UI-TARS-7B w/ \textbf{Ours} & 7B w/ 8B & \textbf{75.3} & \textbf{88.2} & \textbf{65.3}\\
\bottomrule
\end{tabular}
}
\vspace{-1mm}
\caption{\textbf{Step success rate (\%) on AndroidControl-High, GUI-Odyssey, and Multimodal-Mind2Web.} 
\emph{Step SR} reflects the proportion of correctly executed steps, where both the predicted action and its arguments (e.g., click coordinates) match the ground truth. 
Results on Multimodal-Mind2Web are averaged over its cross-task, cross-website, and cross-domain evaluation splits.}
\label{tab:gui-offline}
\end{table*}

\noindent
\textbf{Baselines.} 
We compare \name{} with a broad range of state-of-the-art multimodal agents, covering three major categories. 
\textit{(1) Proprietary models} include o3 and OpenAI CUA-o3~\citep{openai2025o3}, GPT-4o, GPT-4.1~\citep{openai2025o3}, GPT-5~\citep{openai2025gpt5}, Claude 4/4.5 Sonnet and Claude Computer-Use~\citep{claude4}, Seed 1.5-VL~\citep{guo2025seed1}, and UI-TARS-1.5/2~\citep{qin2025ui,wang2025ui}.  
\textit{(2) Agent systems with proprietary models} combine open-source backbones with closed-source planners or reasoning modules, including Jedi-7B~\citep{xie2025scaling}, Agent S2/S2.5~\citep{Agent-S2}, GTA1-7B/32B~\citep{yang2025gta1}, UI-TARS-1.5-7B w/ GPT-4.1, and Qwen3-VL-32B-Thinking w/ GPT-4.1~\citep{yang2025qwen3}.  
\textit{(3) Open-source models} include OS-Atlas~\citep{wu2024atlas}, GUI-R1~\citep{luo2025gui}, Qwen2.5-VL and Qwen3-VL series~\citep{bai2025qwen2,yang2025qwen3}, OpenCUA~\citep{wang2025opencua}, UI-TARS variants~\citep{qin2025ui}, LLAVA-NeXT~\citep{liu2024llavanext}, DeepSeek-VL2~\citep{wu2024deepseek}, and T3-Agent~\citep{gao2024multi}.  
Results for all baselines are mainly taken from their official reports. For our methods, we report the mean performance over $3$ independent runs.

\subsection{Main Results on GUI Environments}

Table~\ref{tab:gui-online} presents results on the online benchmarks, \textit{AndroidWorld} and \textit{OSWorld-Verified}. 
\name{} achieves substantial gains over its grounding models and reaches performance comparable to proprietary GPT-4.1 planners, highlighting the strength of its trajectory-level anticipatory reasoning for long-horizon GUI control. 
Specifically, our method improves the success rate of UI-TARS-1.5-7B from $27.4\%$ to $30.9\%$ on OSWorld-Verified, and boosts Qwen3-VL-32B-Thinking from $35.6\%$ to $41.2\%$, corresponding to relative gains of $12.8\%$ and $15.7\%$, respectively. 
These results demonstrate that anticipatory planning substantially enhances stability and task success across mobile and desktop platforms, establishing new state-of-the-art results among open-source models of comparable size.

As shown in Table~\ref{tab:gui-offline}, our model exhibits strong high-level task planning ability across offline GUI benchmarks. 
Built entirely on open-source backbones, it achieves performance on par with GPT-4.1–based proprietary planners 
Compared with R1-style models trained under distinct training objectives, such as GUI-R1 and InfiGUI-R1, our method delivers substantially stronger results on high-level task execution, exceeding them by more than 40\% on AndroidControl-High and setting a new state of the art among open-source GUI agents. 
These gains underscore the advantage of trajectory-aware reasoning, which enables the model to accurately translate complex, high-level task instructions into fine-grained action instructions, achieving far more reliable execution than reactive agents in compositional GUI environments.

\subsection{Main Results on General Tool-use Scenarios}
\begin{table*}[t]
\centering
\small
\setlength{\tabcolsep}{6pt}
\begin{tabular}{l c cccc cccc}
\toprule
\multirow{2}{*}{\textbf{Agent Model}} &
\multirow{2}{*}{\textbf{Params}} &
\multicolumn{4}{c}{\textbf{GAIA}} &
\multicolumn{3}{c}{\textbf{GTA}} \\
\cmidrule(lr){3-6} \cmidrule(lr){7-9}
& & \textbf{AnsAcc} & \textbf{Level 1} & \textbf{Level 2} & \textbf{Level 3}
  & \textbf{AnsAcc} & \textbf{ToolAcc} & \textbf{CodeExec} \\
\midrule
\rowcolor{cvprgreen!10}
\multicolumn{9}{l}{\textit{Proprietary Models}} \\
GPT-4o~\citep{hurst2024gpt} & -   & 33.4 & 47.1 & 31.4 & 11.5 & 57.1 & 63.4 & 95.1 \\
GPT-4.1~\citep{hurst2024gpt} & -   & 50.3 & 58.5 & 50.0 & 34.6 & 58.4 & 65.1 & 94.3 \\
GPT-5~\citep{openai2025gpt5} & - & 59.3   & 67.9 & 58.1 & 46.1 & 60.9 & 68.3 & 98.7 \\
\midrule
\rowcolor{cvprgreen!10}
\multicolumn{9}{l}{\textit{Open-Source Models}} \\
LLAVA-NeXT-8B~\citep{liu2024llavanext} & 8B & 3.6   &  9.4 & 1.2 & 0.0 & 14.1  & 14.9 &  25.1\\
DeepSeek-VL2~\citep{wu2024deepseek} & 72B & 14.2   & 19.3 & 12.4 & 10.3 & 23.2 & 49.4 & 57.2 \\
Qwen2.5-VL-7B~\citep{bai2025qwen2} & 7B & 10.3   & 16.9 &  9.3 & 0.0 & 44.2 & 50.6 &   69.1 \\
Qwen2.5-VL-14B~\citep{bai2025qwen2} & 14B & 15.2   & 24.5 &  11.6 & 3.8 & 46.8  & 55.4 & 69.8 \\
Qwen3-VL-8B~\citep{bai2025qwen2} & 8B & 31.5   & 46.2 & 27.6 & 16.3 & 49.2 & 56.8 & 74.2\\
T3-Agent~\citep{gao2024multi}      & 7B  &  16.9 &  26.4 & 15.1 & 3.8 & 53.8 & 64.6 & 84.3 \\
\midrule
\name & 8B & \textbf{40.2} & \textbf{55.9} & \textbf{35.8} & \textbf{24.4} & \textbf{56.7} & \textbf{65.7} & \textbf{87.4} \\
\bottomrule
\end{tabular}
\vspace{-1mm}
\caption{\textbf{Results on the GAIA and GTA benchmarks.} 
GAIA assesses AI assistants across three difficulty levels, where final answer accuracy (\emph{AnsAcc}) reflects overall tool-usage reasoning correctness. 
GTA further evaluates multimodal tool-use ability using three metrics: \emph{AnsAcc} for answer correctness, \emph{ToolAcc} for accurate tool selection and summarization, and \emph{CodeExec} for the percentage of generated code that executes without errors.}
\label{tab:tool-bench}
\end{table*}

Table~\ref{tab:tool-bench} presents results on the GAIA and GTA benchmarks. 
\name{} demonstrates robust multimodal reasoning and tool-use ability, outperforming GPT-4o on GAIA and achieving the best performance among all open-source models. 
Compared with Qwen3-VL-8B, it attains a notable $+8.7$ improvement in answer accuracy, reflecting stronger reasoning consistency across the three GAIA levels. 
On GTA, \name{} exhibits exceptional tool-execution behavior with particularly high \emph{ToolAcc}, confirming the effectiveness of training with tool-usage trajectories. 
In addition, the second-stage tool-grounded RFT enhances the reliability of generated code, leading to higher \emph{CodeExec} success and more stable answer generation. 
Taken together, the results suggest that \name{}’s trajectory-level anticipatory reasoning yields more reliable tool use and more coherent decision-making, revealing a unified mechanism for grounded multimodal reasoning.

\subsection{Ablations and Discussions}
\begin{figure}
\centering
\small
\setlength{\tabcolsep}{6pt}
\renewcommand{\arraystretch}{1.1}
\begin{tabular}{lccc}
\toprule
\textbf{Setting} & \textbf{AndroidWorld} & \textbf{OSWorld-Verified} & \textbf{GTA} \\
\midrule
w/ Stage~$2$  & 64.8 & 41.2 &  56.7 \\
w/o Stage~$2$ & 57.2 & 36.3 &  50.2 \\
\bottomrule
\end{tabular}
\vspace{-1mm}
\caption{\textbf{Two-stage Training Ablation.} 
Performance (\%) comparison on AndroidWorld, OSWorld-Verified and GTA benchmarks. 
Stage~$2$ provides consistent improvements in both settings.}
\label{tab:twostage}
\end{figure}
\begin{figure*}[t]
    \centering
    \includegraphics[width=0.99\linewidth]{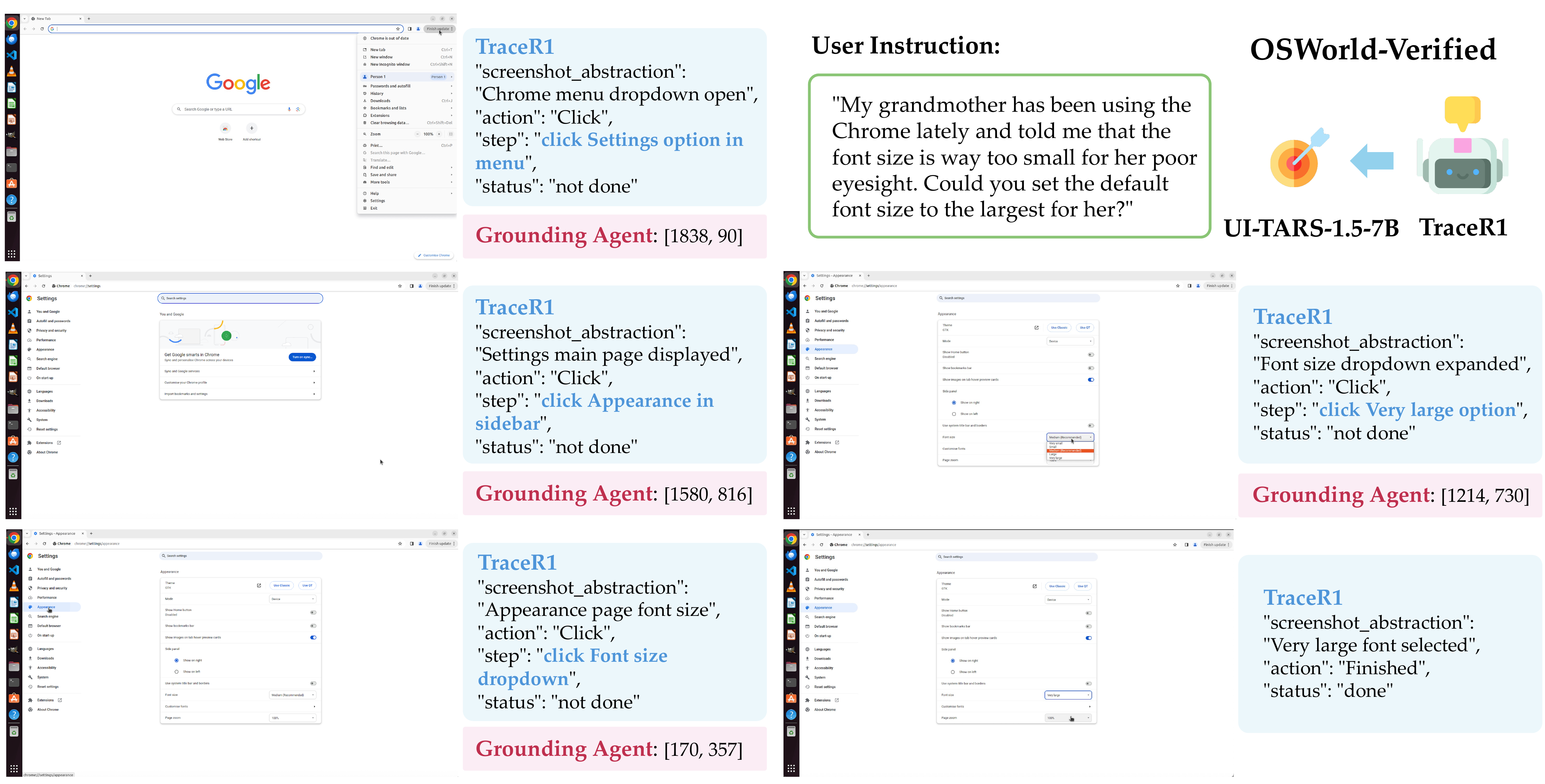}
    \vspace{-1mm}
    \caption{\textbf{Example trajectory:} coordination between TraceR1 (planner) and UI-TARS-1.5-7B (executor) on OSWorld-Verified.}
    \label{fig:case}
\end{figure*}
\begin{figure*}[t]
    \centering
    \begin{subfigure}{0.28\linewidth}
        \includegraphics[width=\linewidth]{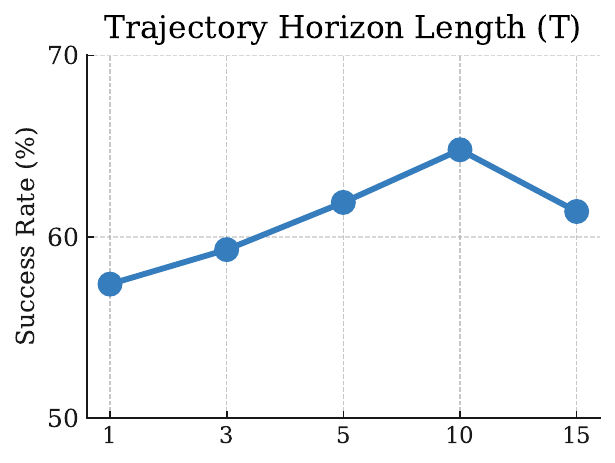}
        \caption{Trajectory horizon length}
        \label{fig:ablation_a}
    \end{subfigure}
    \begin{subfigure}{0.34\linewidth}
        \includegraphics[width=\linewidth]{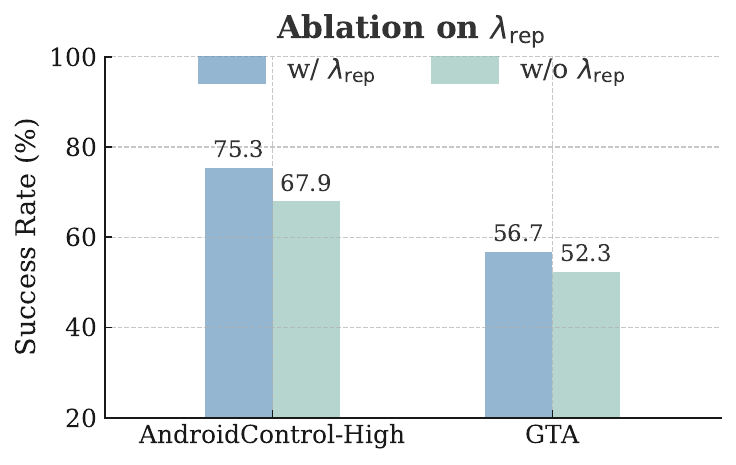}
        \caption{Ablation on repetition penalty $\lambda_{\text{rep}}$}
        \label{fig:ablation_b}
    \end{subfigure}
    \begin{subfigure}{0.34\linewidth}
        \includegraphics[width=\linewidth]{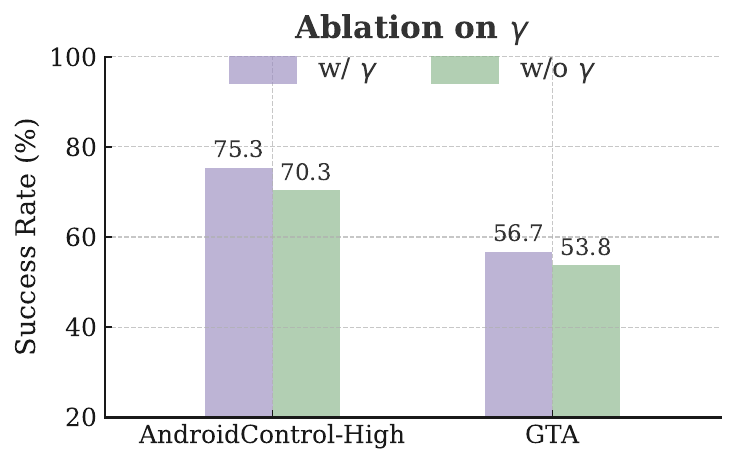}
        \caption{Ablation on temporal discount $\gamma$}
        \label{fig:ablation_c}
    \end{subfigure}
    \vspace{-1mm}
    \caption{\textbf{Ablation results.}
    (a) shows the effect of predictive trajectory horizon length on AndroidWorld.
    (b--c) show the impact of removing $\lambda_{\text{rep}}$ and $\gamma$ during training
    on AndroidControl-High and GTA.}
    \label{fig:ablation_three}
\end{figure*}
\noindent
\textbf{Incorporating execution feedback stabilizes long-horizon planning.}
As shown in Table~\ref{tab:twostage}, removing Stage~2 leads to an average performance drop of roughly 6\%, which demonstrates the importance of grounded execution signals for stable plan generation.
Without this stage, the planner is trained only with abstract trajectory-level rewards and receives no information about whether its predicted actions are actually feasible.
This lack of grounding often produces unstable or overly optimistic plans, such as assuming nonexistent tools or expecting successful executions that never materialize.
Stage~2 provides the model with concrete execution outcomes that serve as corrective signals, enabling it to adjust its predictions and maintain coherent and feasible plans across different environments.

\noindent
\textbf{Balancing prediction horizon.}
We vary the predictive horizon $T$, which controls how many future steps the planner learns to forecast during training.
As shown in Figure~\ref{fig:ablation_a}, increasing 
$T$ initially improves task success, as the model benefits from learning to anticipate delayed outcomes and organize temporally extended plans.
However, beyond a moderate range ($T>10$), performance drops noticeably.
When trained with excessively long horizons, the planner must predict far-future transitions whose uncertainty accumulates quickly, leading to noisy trajectory rewards and unstable credit assignment.
This result suggests that training the model to look several steps ahead is beneficial for long-term reasoning, but forcing it to plan too far into the future overwhelms the learning signal and weakens adaptation during execution.

\noindent
\textbf{Trajectory-aware reward design analysis.}
We further analyze the impact of two key components in our trajectory-level reward: the repetition penalty $\lambda_{\text{rep}}$ and the temporal discount factor $\gamma$.
As shown in Figure~\ref{fig:ablation_b} and~\ref{fig:ablation_c}, removing the repetition penalty ($\lambda_{\text{rep}}=0$) or disabling temporal discounting ($\gamma=1$) consistently degrades performance, confirming that both components are crucial for stabilizing trajectory-aware reinforcement learning.
Without the repetition penalty, the planner exhibits reward-hacking behavior by repeatedly issuing redundant actions, such as re-clicking the same interface element or re-invoking identical tools, in order to inflate short-term rewards without making real task progress.
This behavior undermines both efficiency and causal consistency in long-horizon planning.
The temporal discount factor further mitigates instability by prioritizing near-future correctness while preserving trajectory-level coherence, preventing the planner from overfitting to uncertain distant outcomes and yielding more consistent, goal-directed planning.
\section{Conclusion}
\label{sec:cons}
We presented \textbf{\name}, a two-stage RL framework that endows multimodal agents with anticipatory planning ability across GUI and tool-use environments.
By coupling trajectory-level optimization with grounded execution refinement, \name{} bridges the gap between high-level foresight and low-level precision, achieving substantial gains in planning stability, execution reliability, and generalization.
Our experiments demonstrate that trajectory-aware reasoning can significantly improve the coherence and adaptability of multimodal agents, establishing a scalable recipe for training open models to reason and plan ahead within dynamic, state-changing environments.
More broadly, this work highlights anticipatory trajectory reasoning as a general principle for building agents capable of coherent, temporally extended decision making across heterogeneous interaction modalities.

Although effective, the current approach is limited because short-horizon updates provide local corrections and cannot reshape the agent’s understanding of long-term feasibility or task structure. Future work may explore multi-round or hierarchical planning mechanisms that couple trajectory prediction with updates to memory, internal state, or world models, allowing the agent to revise and consolidate plans. Another promising direction is to extend this paradigm to embodied or hybrid tool-use environments, where successful behavior requires coordinating perception, reasoning, and action across longer time scales. Advances along these lines may yield planning systems that not only anticipate future outcomes but also organize their predictions across multiple levels of abstraction.

{
    \small
    \bibliographystyle{ieeenat_fullname}
    \bibliography{main}
}

\clearpage
\appendix
\definecolor{custom_blue}{RGB}{235,244,253}
\tcbset{
  aibox/.style={
    width=\textwidth,
    top=10pt,
    colback=white,
    colframe=black,
    colbacktitle=black,
    center,
  }
}

\newtcolorbox{graybox}[1][]{colback=lightgray!20, colframe=lightgray!50!black, boxrule=0pt, enhanced, #1}
\newtcolorbox{redbox}[1][]{colback=red!10, colframe=red!50!black, boxrule=0pt, enhanced, #1}
\newtcolorbox{greenbox}[1][]{colback=green!10, colframe=green!50!black, boxrule=0pt, enhanced, #1}
\newtcolorbox{magentabox}[1][]{colback=magenta!10, colframe=magenta!50!black, boxrule=0pt, enhanced, #1}
\definecolor{lightgreen}{RGB}{144, 238, 144} 

\newtcolorbox{AIbox}[2][]{aibox,title=#2,#1}
\tcbset{
  aiboxsmall/.style={
    width=0.62\textwidth,
    top=10pt,
    colback=white,
    colframe=black,
    colbacktitle=black,
    enhanced,
    top,
    attach boxed title to top left={yshift=-0.1in,xshift=0.15in},
    boxed title style={boxrule=0pt,colframe=white,},
  }
}   
\definecolor{commentcolor}{RGB}{34,139,34} 
\newcommand{\myalgorithm}{%
\begingroup
\removelatexerror
\begin{algorithm*}[H]
      \caption{\modelname PyTorch pseudocode.}
      \label{alg:pseudocode}
      \scriptsize
          \Comment{
          $\mathbf{H}_0$: Input embeddings for LLM (Original inputs args for traditional LMM); \\
          $vis\_pos$: the location of visual tokens; \\
          $\mathbf{X}$, $\mathbf{X^{stack}}$: Original visual tokens, Extra high-resolution visual token list; \\
          $l_{start}$, $n$: Index of starting layer, and layer interval for stacking.
          }
  \Function{forward($\mathbf{H}_0$, $\mathbf{X^{stack}}$, $l_{start}$, $n$, $vis\_pos$)}{
     \var{$\mathbf{H}$ = $\mathbf{H}_0$}
     
     \For{($idx$, \var{TransformerLayer)} in enumerate(\var{self.layers})}{
        \Comment{\modelname:}
        \If{$idx$ >= $l_{start}$ \& $(idx - l_{start}) \% n == 0$}
        {
            $\mathbf{H}[vis\_pos]$ += $\mathbf{X^{stack}}[(idx - l_{start})//n]$
        }

        \Comment{Original Transformer:}
        $\mathbf{H}$ = \var{TransformerLayer}($\mathbf{H}$)
     }
  }
\end{algorithm*}
\endgroup}

\setcounter{page}{1}
\maketitlesupplementary

\section{Implementation Details}
\label{app:train}
\subsection{Training Hyperparameters}
Table~\ref{tab:grpo_hparams} summarizes the GRPO and optimization
hyperparameters used in our experiments. Both Stage~1 (trajectory-level
optimization) and Stage~2 (grounded fine-tuning) share identical training
configurations; the only difference lies in their reward definitions.

\begin{table*}[t]
\centering
\footnotesize
\setlength{\tabcolsep}{8pt}
\resizebox{0.6\textwidth}{!}{
\begin{tabular}{llc}
\toprule
\textbf{Category} & \textbf{Hyperparameter} & \textbf{Value} \\
\midrule

\multirow{10}{*}{\textbf{Actor Optimization}} 
& Learning rate & $1\times 10^{-6}$ \\
& Optimizer & AdamW (bf16) \\
& Weight decay & 0.01 \\
& Warmup ratio & 0 \\
& Training steps & 143 \\
& Max grad norm & 1.0 \\
& Global batch size & 128 \\
\midrule

\multirow{10}{*}{\textbf{GRPO / RL Parameters}} 
& Advantage estimator & GRPO \\
& Discount ($\gamma$) & 0.8 \\
& GAE $\lambda$ & 1.0 \\
& KL type & fixed \\
& KL target & 0.1 \\
& KL coef & 0.01 \\
& KL penalty & low\_var\_kl \\
& Clip ratio (low) & 0.2 \\
& Clip ratio (high) & 0.3 \\
& Clip ratio (dual) & 3.0 \\
\midrule

\multirow{8}{*}{\textbf{Rollout Generation}} 
& Engine & vLLM \\
& Number of rollouts (n) & 5 \\
& Rollout batch size & 512 \\
& Temperature & 1.0 \\
& Top-$p$ & 0.99 \\
& Tensor parallel size & 2 \\
& Max batched tokens & 8192 \\
& Response length & 2048 \\
\midrule

\multirow{3}{*}{\textbf{Reward Parameters}} 
& $\lambda_{\text{align}}$ & 0.8\\
& $\lambda_{\text{rep}}$ & 0.1\\
& $\lambda_{\text{format}}$ & 0.1\\
\bottomrule
\end{tabular}}
\caption{\textbf{GRPO training hyperparameters used for both Stage~1 and Stage~2.} 
All reinforcement learning stages share the same GRPO and optimization settings. }
\label{tab:grpo_hparams}
\end{table*}

\subsection{Reward Formulation Details}
\label{app:reward_details}

Based on the provided implementation, the reward function in Stage 1 aims to align the predicted trajectory skeleton (action types and status) with the ground truth, while strictly enforcing output formatting. The total reward $R$ for a predicted sample is a weighted sum of the accuracy reward $R_{\text{acc}}$ and the format reward $R_{\text{fmt}}$:
\begin{equation}
    R = (1 - \lambda_{\text{fmt}}) \cdot R_{\text{acc}} + \lambda_{\text{fmt}} \cdot R_{\text{fmt}}
\end{equation}
where we set $\lambda_{\text{fmt}} = 0.1$.

\paragraph{Format Reward ($R_{\text{fmt}}$).} 
To ensure the model generates parseable actions, we check for the presence of specific XML tags (e.g., \texttt{<think>}, \texttt{<answer>}) and JSON keys. For a generated response containing $N$ steps, the format reward is the ratio of valid steps:
\begin{equation}
    R_{\text{fmt}} = \frac{1}{N} \sum_{i=1}^{N} \mathbb{1}[\text{Valid}(\hat{s}_i)]
\end{equation}
A step $\hat{s}_i$ is considered valid only if it strictly contains the required keys: \texttt{"screenshot\_abstraction"}, \texttt{"action"} (with \texttt{"action\_type"}), and \texttt{"status"}.

\paragraph{Trajectory Alignment Accuracy ($R_{\text{acc}}$).}
Unlike standard exact matching, our trajectory-level reward focuses on the correctness of the \textit{plan sequence} (Action Types). Let $\mathcal{A} = [\hat{a}_1, \dots, \hat{a}_N]$ be the predicted action sequence and $\mathcal{A}^* = [a^*_1, \dots, a^*_M]$ be the ground truth. We parse only the \texttt{action\_type} and \texttt{status} fields for alignment.

\noindent \textit{(1) Greedy Alignment with Position Penalty.} 
We compute the best alignment between $\mathcal{A}$ and $\mathcal{A}^*$. If lengths differ, we employ a greedy matching strategy. A predicted action $\hat{a}_i$ matches a ground truth action $a^*_j$ if:
\begin{equation}
    \text{sim}(\hat{a}_i, a^*_j) = \mathbb{1}[\hat{a}_i.\text{type} = a^*_j.\text{type}]
\end{equation}
To encourage temporal consistency, we apply a position penalty $P_{\text{pos}} = |i - j| \times 0.1$. A match is accepted only if the adjusted score $(\text{sim} - P_{\text{pos}}) > 0.5$.

\noindent \textit{(2) Discounted Score.}
For the set of aligned pairs $\mathcal{M}$, the base alignment score is calculated using a discount factor $\gamma=0.8$:
\begin{equation}
    S_{\text{align}} = \sum_{(i,j) \in \mathcal{M}} \gamma^i \cdot \text{sim}(\hat{a}_i, a^*_j)
\end{equation}
We subtract a coverage penalty of $0.15$ for every unmatched action in both prediction and ground truth. The score is normalized by the maximum possible discounted return of the reference trajectory.

\noindent \textit{(3) Repetition Penalty.}
Finally, we penalize stuck loops (e.g., repeatedly predicting "click" without state change). If three consecutive actions have the same type, it counts as a repetition.
\begin{equation}
    R_{\text{acc}} = \text{Clip}_{[0,1]} \left( \bar{S}_{\text{align}} - 0.1 \times N_{\text{repetitions}} \right)
\end{equation}
where $\bar{S}_{\text{align}}$ is the normalized alignment score. This design encourages the agent to follow the correct high-level plan before refining parameters in Stage 2.

\end{document}